# Probabilistic Similarity Logic

**Matthias Bröcheler**
Computer Science Dept.
University of Maryland
College Park, MD 20740

**Lilyana Mihalkova**
Computer Science Dept.
University of Maryland
College Park, MD 20740

**Lise Getoor**
Computer Science Dept.
University of Maryland
College Park, MD 20740

## Abstract

Many machine learning applications require the ability to learn from and reason about noisy multi-relational data. To address this, several effective representations have been developed that provide both a language for expressing the structural regularities of a domain, and principled support for probabilistic inference. In addition to these two aspects, however, many applications also involve a third aspect–the need to reason about similarities–which has not been directly supported in existing frameworks. This paper introduces probabilistic similarity logic (PSL), a general-purpose framework for joint reasoning about similarity in relational domains that incorporates probabilistic reasoning about similarities and relational structure in a principled way. PSL can integrate any existing domain-specific similarity measures and also supports reasoning about similarities between sets of entities. We provide efficient inference and learning techniques for PSL and demonstrate its effectiveness both in common relational tasks and in settings that require reasoning about similarity.

## 1 Introduction

A variety of machine learning applications require the ability to learn from and reason about noisy, or uncertain, multi-relational data. This has motivated the fields of statistical relational learning (SRL) and multi-relational data mining, which have made significant progress on developing effective representations that incorporate in a unified framework two central aspects of modeling in multi-relational domains—on the one hand, these representations provide a language for expressing the structural regularities present in a domain, and on the other hand, they provide principled support for probabilistic inference, e.g., [12, 8].

In addition to relational structure and probabilistic dependencies, many applications of interest also involve a third aspect—the need to reason about similarities—which has not been directly supported in existing SRL frameworks.

This paper introduces *probabilistic similarity logic* (PSL), a general-purpose framework for joint reasoning about similarity in relational domains. PSL uses annotated rules to capture the dependency structure of the domain, based on which it builds a joint probabilistic model over all similarity decision atoms. PSL embodies the following novel characteristics. First, PSL provides a unified framework in which probabilistic reasoning about relational structure is seamlessly incorporated with reasoning about similarities. A direct consequence of this is that PSL can integrate any existing similarity measures, thereby extending their applicability to a relational context. It is important to distinguish between similarity itself and the uncertainty in similarity propagation - the latter justifying our probabilistic approach. Second, as an extension to its treatment of similarity, PSL supports reasoning about similarity between *sets* of entities, defined by a given relation. Third, PSL is efficient because it casts inference as a cone program and uses a relational database for data management, which allows it to take advantage of efficient querying techniques developed in the database community.

**Motivating Examples:** To illustrate the diversity of relational settings that require reasoning about similarity, we next describe two distinct applications. The first motivating application is a Wikipedia-like environment in which a set of hyperlinked documents are being edited by a set of interacting users. One task in this setting, that is of interest in information retrieval applications, is to automatically infer similarities between the documents based both on their content and on the relational structure of the domain. Alternatively, one may be interested in identifying similar users, e.g. in collaborative filtering. Each of these problems can be approached in isolation by comparing pairs of entities based on their attributes. For example, for document similarity, one can take advantage of the extensive literature on similarity metrics, e.g., [2]. However, due to the relational structure of this problem, user similarity and document similarity are closely entangled. For instance, users who edit similar documents are likely to be similar. Conversely, documents edited by similar users are likely to be similar. Furthermore, the relational structure of the domain can also be incorporated into this reasoning to state, for example, that two documents are likely similar if they have been edited by users who interact frequently. Set constructs allow for significant modeling flexibility in this domain. For example, let $U_1$ and $U_2$ stand for two users and $\{U.edited\}$ represents the set of all documents edited by $U$; then, one can write $U_1 \cong U_2 \Rightarrow \{U_1.edited\} \cong \{U_2.edited\}$ to state that if $U_1$ is similar to $U_2$ we conclude that the sets of documents edited by $U_1$ and $U_2$ are similar.

As a second motivating application, consider the task of ontology alignment. An ontology is a formal specification of a set of concepts and the different relationships that exist among them, usually forming a concept hierarchy. The goal of ontology alignment is, given two ontologies $O_1$ and $O_2$ that may use different vocabularies to describe the same, or similar, concepts, to find a matching between the concepts and relationships in $O_1$ and $O_2$, e.g., [6, 10]. Because frequently no exact match exists, one needs to reason about degrees of similarity between concepts and relations, while at the same time incorporating this reasoning into a relational framework. For example, one can exploit regularities such as that two concepts are similar if their subconcepts or parent concepts are similar.

Several other applications could also benefit from the ability to incorporate similarities into a relational framework. For example, in computer vision, the similarity of two images can be based both on domain-specific similarity measures and on relational structure within the image; in bioinformatics, one may predict the function of a protein based on its similarity to other proteins, inferred from its properties, and protein-protein interactions.

The paper first introduces PSL in Section 2; then we evaluate PSL on the problems of inferring document similarity and ontology alignment. We evaluate the importance of PSL's elements and demonstrate how PSL can obtain state-of-the-art performance in ontology alignment by incorporating domain-specific similarity measures. Related work is discussed in Section 4.

## 2 PSL

PSL specifies how similarities propagate through the relational structure using annotated rules. PSL represents a family of languages, defined by particular user choices. In the following general description, we discuss the choice points and explain the specific choices made in this paper.

### 2.1 Syntax of PSL

For maximum expressivity, PSL rules can be written in first-order logic (FOL); in addition, we support an object-oriented (OO) short-hand that is more succinct in many cases. In particular, let $X$ and $Y$ be variables representing entities in the domain. Entities are typed, and $T_X$ is the type of $X$, e.g., person, document, etc. Each entity $X$ has a set of attributes $\mathcal{A}(T_X)$ and may participate in a set of relations $\mathcal{R}(T_X)$. Using FOL, $a(X,V)$ asserts that $X$ has attribute $a \in \mathcal{A}(T_X)$ with value $V$. Using OO, $X.a$ returns the value of attribute $a$. Analogously, for a relation $r \in \mathcal{R}(T_X)$, in FOL $r(X,Y)$ indicates that $X$ and $Y$ are related via $r$, and in OO, $X.r$ refers to any one entity related to $X$ via $r$. PSL can also represent sets of entities. If $X.edited$ refers to a document $X$ edited, then $\{X.edited\}$ refers to the set of *all* documents edited by $X$. The same statement can be expressed algebraically as $\{Y|edited(X,Y)\}$. While the latter notation is cumbersome, it allows for sets defined by $n$-ary relations, where $n > 2$, i.e., $\{Y|edited(X,Y,T)\}$ may refer to all documents $Y$ that were edited by user $X$ at time $T$. Although both types of syntax are supported by PSL, for clarity, we will use primarily the OO syntax.

PSL also supports statements about the similarity between two entities or relation-defined sets of entities. We can write statements such as $X.text \stackrel{s_1}{\cong} Y.text$, meaning that the textual content of documents $X$ and $Y$ is similar according to measure $s_1$; $X.edited \stackrel{s_2}{\cong} Y.edited$, meaning that one of the documents edited by $X$ is similar to one of the documents edited by $Y$; and $\{X.edited\} \stackrel{s_3}{\cong} \{Y.edited\}$, meaning that *the entire set* of documents edited by $X$ is similar to the set of $Y$'s edits. The functions $s_1$, $s_2$, and $s_3$ can be *arbitrary* functions of a pair of the appropriate types of entities, or sets of entities, whose domain is $[0,1]$.

Similarity statements combined in logical rules form a PSL program. A PSL program may contain three types of rules: soft rules, each of which has a weight that determines the relative importance of the rule, as discussed in Section 2.2; hard constraints, which are always required to hold; and exclusivity constraints. Hard constraints can be viewed as rules with infinite weight but are maintained separately in PSL to enforce them throughout inference. An exclusivity constraint on a relation $r$ and entity $X$ states that $X$ can be related to at most one entity via $r$.

Table 1 shows an example PSL program for our Wikipedia domain. The first rule states that if two documents have similar text, then they are similar; the second rule states that two documents are similar if the sets of their editors are similar; the third rule states that two documents are similar if the sets of their first- and second-order neighbors in the hyperlink graph are similar; the fourth rule encodes transitivity of similarity and is a hard constraint. The similarity function $s_n$ is based on attributes and $s_{\{\}}$ is defined below.

However, unlike in binary logic, here the conjunction and implication operators need to combine similarities, which are real numbers in $[0,1]$. To emphasize this distinction, we have placed a tilde over these operators in Table 1. Moreover, we would also like to have "soft" versions of disjunction and negation that behave as their counterparts from binary logic so that we are able to manipulate PSL rules as in logic. For example, the fourth rule could be rewritten as a disjunction as follows: $\tilde{\neg}(A \stackrel{s_p}{=} B) \tilde{\vee} \tilde{\neg}(B \stackrel{s_p}{=} C) \tilde{\vee} A \stackrel{s_p}{=} C$. One set of such truth-combining operators that generalize their Boolean counterparts is provided by t-norms and their corresponding t-conorms [14]. In PSL, any t-norm/t-conorm pair may be used. We used the Lukasiewicz t-(co)norm, defined as follows:

$$a \tilde{\wedge} b = \max\{0, a + b - 1\} \quad (1)$$
$$a \tilde{\vee} b = \min\{a + b, 1\} \quad (2)$$
$$\tilde{\neg} a = 1 - a \quad (3)$$

| |
|---|
| $w_1 : A.text \stackrel{s_n}{=} B.text \Rightarrow A \cong B$ |
| $w_2 : \{A.editor\} \stackrel{s_{\{\}}}{=} \{B.editor\} \Rightarrow A \cong B$ |
| $w_3 : \{A.linksTo\} \cup \{A.linksTo.linksTo\}$ $\stackrel{s_{\{\}}}{=} \{B.linksTo\} \cup \{B.linksTo.linksTo\} \Rightarrow A \cong B$ |
| Hard : $A \cong B \tilde{\wedge} B \cong C \Rightarrow A \cong C$ |

Table 1: Example PSL program.

Above, $a, b \in [0, 1]$ can be similarities or Boolean truth values. The Lukasiewicz t-norm is appealing because it is linear in the values being combined, and because unlike, for example, the product t-norm, which defines $a \tilde{\wedge} b = ab$, the Lukasiewicz t-norm leads to sparser grounded PSL programs because the $\tilde{\wedge}$ operator evaluates to 0 on all $a, b$ for which $a + b < 1$. On the other hand, the product t-norm may be more appropriate in domains in which it is important to model longer-range dependencies.

For computational efficiency, we currently require sets to be fully observed, e.g., all groundings of the *edited* relation in $\{A.edited\}$ must be provided as evidence. However, as an essential feature, PSL supports set similarity functions that combine the results of reasoning over similarities between individual members of the sets. For example, in $\{A.edited\} \stackrel{s_{\{\}}}{=} \{B.edited\}$, $s_{\{\}}$ can be defined as

$$s_{\{\}} = \frac{\sum_{i \in \{A.edited\}} \sum_{j \in \{B.edited\}} s_p(i, j)}{|\{A.edited\}| + |\{B.edited\}|},$$

where $s_p(i, j)$ is a function of the similarity of two individuals. The values of $s_p$ can be inferred from the data and are *not* required to be part of the evidence.

### 2.2 Semantics of PSL

A PSL program defines a probability distribution over similarities between entities or sets of entities in a domain. In the following discussion, we require that all weights be positive. This requirement does not detract from the generality of PSL because any negative weight can be made positive by negating the corresponding rule.

Given a domain $\mathcal{D}$, each **grounding** of each PSL rule $R$ represents an instantiation of all variables in $R$ by replacing them with entities from $\mathcal{D}$. For each rule, all possible groundings are generated. For example, let $R$ be:

$$\{A.editor\} \stackrel{s_1}{=} \{B.editor\} \tilde{\wedge} A.text \stackrel{s_2}{=} B.text \Rightarrow A \stackrel{s_3}{=} B$$

Suppose $\mathcal{D}$ contains the entities $doc1$ and $doc2$. Then, there are 4 unique **groundings** of $R$, corresponding to the possible ways of replacing $A$ and $B$ with $doc1$ and $doc2$, and in each grounding we expand the *editor* relation and the *text* attribute of each participating entity. A statement of the form $a \stackrel{s_i}{=} b$, where $a$ and $b$ are entities or sets of entities, is called a **ground proposition**. Each ground proposition represents a statement about particular entities and can be assigned a value by the similarity function $s_i$. Let $\mathcal{G}$ be the set of all ground propositions with the entities in $\mathcal{D}$. Let $I(\mathcal{G})$ be an **interpretation**, i.e., a particular truth assignment to the elements in $\mathcal{G}$, such that for each $g \in \mathcal{G}$, its truth value is a real number between 0 and 1, i.e., $I(g) \in [0, 1]$. The probability of an interpretation $I(\mathcal{G})$, according to a PSL program $\mathcal{P}$ is given by the following expression:

$$\mathbb{P}(I(\mathcal{G})) = \frac{1}{Z} \exp(-d_\delta(\mathcal{P}, I)) \quad (4)$$

Above, $Z = \int_{I'} \exp(-d_\delta(\mathcal{P}, I'))$ is the familiar normalizing constant that integrates over possible real-valued truth assignments. In the exponent, $d_\delta(\mathcal{P}, I)$ is the **distance from satisfaction** function. First, we define the distance from satisfaction of a single grounding $G_R$ of rule $R \in \mathcal{P}$.

**Definition 1** *The distance from satisfaction of a single ground rule $G_R$, according to a particular interpretation $I$, is $d(G_R, I) = 1 - I(G_R)$.*

Intuitively, the closer the value of a particular ground rule is to 1, the closer it is to being satisfied, and the smaller its distance from satisfaction. We now extend this definition to the distance from satisfaction of a PSL program $\mathcal{P}$. This is the distance from satisfaction of all possible groundings of the rules in $\mathcal{P}$, weighted by the weight of each rule.

**Definition 2** *Let $\mathcal{P}$ be a PSL program containing $n$ rules and let $I$ be an interpretation. If $I$ violates any hard or exclusivity constraints in $\mathcal{P}$, $\mathcal{P}$'s distance from satisfaction is $\infty$. Otherwise, for each rule $R_k \in \mathcal{P}$, let $V_k(I)$ be the vector containing the distances from satisfaction of all groundings of $R_k$ in $I$, i.e., $V_k = [d(G_{R_k}^1, I) \ldots d(G_{R_k}^{n_{R_k}}, I)]^T$, where $n_{R_k}$ is the number of groundings of $R_k$. Let $V(I)$ be the vector formed by stacking all of the $V_k$-s after multiplying them with the corresponding weight $w_k$: $V(I) = [w_1 V_1(I) \ldots w_n V_n(I)]^T$. Let $\delta$ be an arbitrary distance metric. Then, $d_\delta(I, \mathbf{P}) = \delta(V(I), \mathbf{0})$.*

The distance metric $\delta$ presents another choice point by which members in the PSL family of languages are identified. For example, if we use the $L_1$-norm distance $\delta_{L_1}(\mathbf{x}, \mathbf{y}) = \|\mathbf{x} - \mathbf{y}\|_1$, we obtain the log-linear representation, commonly used in SRL and graphical models. Alternatively, we could use the squared $L_2$-norm distance $\delta_{L_2}(\mathbf{x}, \mathbf{y}) = \|\mathbf{x} - \mathbf{y}\|_2^2$, thus making the penalty for not satisfying a rule a faster-growing function of the distance from satisfaction.

## 2.3 The Importance of Similarity

The ability to reason about similarities in a relational framework is a central novel property of PSL. Here we motivate its importance. The most immediate advantage of reasoning about similarity is that, in PSL, the numerous well-understood domain-specific similarity measures that exist in the literature can be easily brought to bear in a relational context. A further advantage results from the interplay of relational structure and similarity; namely, PSL supports reasoning about similarity not only between the attributes of two entities $X$ and $Y$, but also between the respective sets of entities related to $X$ and $Y$, e.g., the sets of entities related to $X$ and $Y$ via the *editor* relation.

Because support for set similarity is such an important aspect of PSL, we consider it further by contrasting setFree-PSL, in which set similarity is not allowed, to the complete PSL. Suppose we would like to reason about the similarity between two documents based on the editors they have in common. This is expressed in setFree-PSL as:

$$A.editor \stackrel{s_s}{\cong} B.editor \tilde{\Rightarrow} A \cong B \quad (5)$$

The main issue with this rule is that the number of its groundings that are active during inference depends on the *absolute* number of editors that $A$ and $B$ have in common. Consider what happens as a result. Let $a_1$ and $b_1$ be two documents, each having $n$ editors with perfect overlap between their editor sets; and let $a_2$ and $b_2$ be two documents each having $m$ editors, $m \gg n$, such that they have $n$ editors in common. Then, all else being equal, the penalty for not inferring that $a_1$ and $b_1$ are similar is equal to the penalty for not inferring that $a_2$ and $b_2$ are similar, although in the former case we have much stronger evidence of the similarity of the two documents. A related issue is that to maintain the relative importance of rules constant across domains, when rules such as the above are present in the model, their weight needs to depend on the sizes of the relations. For example, if a weight for the above rule is learned in one data set and used for prediction in another one in which documents have larger numbers of editors, all else being equal, the relative importance of that rule will increase simply because it will have more active groundings during inference. These issues are completely resolved by the introduction of sets. For example, in PSL we can write:

$$\{A.editor\} \stackrel{s_{\{\}}}{\cong} \{B.editor\} \tilde{\Rightarrow} A \cong B \quad (6)$$

This rule now has a single grounding, and the strength of the evidence on the left equals the amount of overlap between the two sets.

Sets are also beneficial when they appear in the consequent of a rule. Consider the difference between the following two rules that relate the similarity of two concepts in an ontology to the similarity of their sub-concepts:

$$C_1 \stackrel{s_1}{\cong} C_2 \tilde{\Rightarrow} C_1.subconcept \stackrel{s_1}{\cong} C_2.subconcept \quad (7)$$

$$C_1 \stackrel{s_1}{\cong} C_2 \tilde{\Rightarrow} \{C_1.subconcept\} \stackrel{s_{\{\}}}{\cong} \{C_2.subconcept\} \quad (8)$$

For two similar concepts that each have $n$ sub-concepts, the first rule will have at most $n$ true groundings out of $n^2$ possible ones, even if their sub-concepts align perfectly, while the second rule correctly captures the intended meaning.

As a further benefit of sets, using rules such as (6) and (8) instead of (5) and (7) leads to fewer groundings per rule. Specifically, if rule (8) is used, then there will be a single grounding for each pair of concepts $C_1, C_2$; on the other hand, if rule (7) is used, there will be $k^2$ groundings for each pair $C_1, C_2$, where $k$ is the maximum relation size.

## 2.4 Inference

It is frequently necessary to perform maximum *a posteriori* inference (also called MPE inference) to infer the most likely values for a set of propositions, given observed values for the remaining (evidence) propositions. For example, in an ontology alignment task, we would like to predict the best matching of concepts from one of the ontologies with concepts from the other one. More formally, we split the set of propositions into two subsets: let $\mathbf{y}$ be the set of propositions with unknown values and let $\mathbf{x}$ be the set of evidence propositions with values in $I(\mathbf{x})$. Then the task is to find a truth assignment $I_{MAP}(\mathbf{y})$ that is most likely according to the PSL program, given the evidence:

$$I_{MAP}(\mathbf{y}) = \arg\max_{I(\mathbf{y})} P(I(\mathbf{y})|I(\mathbf{x})) \quad (9)$$

$$= \arg\max_{I(\mathbf{y})} \frac{1}{Z} \exp\left(-d_\delta((I(\mathbf{y}), I(\mathbf{x})), \mathcal{P})\right) \quad (10)$$

$$= \arg\max_{I(\mathbf{y})} \left(-d_\delta((I(\mathbf{y}), I(\mathbf{x})), \mathcal{P})\right) \quad (11)$$

As can be seen from Definition 2, to evaluate $d_\delta$, we need to form all groundings of the rules in $\mathcal{P}$ with the entities in the domain. To limit the number of grounded rules that are active during inference, we take advantage of the fact that only grounded rules that evaluate to strictly less than 1 need to be considered. Thus, grounded rules that have value 1 given the evidence $I(\mathbf{x})$ can be excluded from consideration because their value does not depend on assignments to the propositions in $\mathbf{y}$. Furthermore, rather than performing inference over all remaining grounded rules at once, we employ a lazy grounding technique, whereby only grounded rules whose value becomes smaller than 1 at some point during inference are included in the inference problem. Such rules are called **activated** in Alg. 1, which describes MAP inference in PSL. This algorithm works by transforming the grounded PSL program into a second-order cone program (SOCP) [16] (line 4) whose solution gives an assignment to the propositions in $y$ (line 5). Necessary conditions are given in Theorem 1 below. Any rules that have not achieved their maximal possible value of 1 in the current solution, are added to the set of active ground rules (lines 6-10) and the process repeats for as long as new rules are activated. In generating the SOCP (line 5) we introduce one variable $v_a$ for each ground atom $a$. For each ground rule $G_R$ we add an auxiliary variable $v_R$ and the constraint:

$$v_R \geq d(G_R, I'),$$

where $d(G_R, I')$ is as given in Definition 1, and $I'$ is the interpretation in which the truth values of all atoms $a$ are given by their respective variables $v_a$. In words, this constraint specifies that $v_R$ must be greater than or equal to the distance from satisfaction of the rule $G_R$ where the truth values of all atoms $a$ are replaced by their respective variables $v_a$. Moreover, we add all hard and exclusivity domain constraints in the form of constraints on the atom variables $v_a$. Finally, we build the conic objective function, $\delta(V(I'), 0)$, from the (fixed) weights and auxiliary variables $v_R$ according to Definition 2. To solve the SOCP, one can use any available technique [1].

Alg. 1 is in essence equivalent to cutting plane inference (CPI) [20], except that, unlike CPI, PSL programs are continuous constrained numeric optimization programs. Moreover, in the subset of PSL studied here, inference is polynomial in the number of grounded rules activated by Alg. 1, as stated in Theorem 1, which relies on the standard complexity result for SOCP [16] by showing the equivalence of PSL inference to a certain numeric problem outlined above.

**Theorem 1** *Let $R$ be the set of grounded rules activated in Alg. 1. Then, under the choices made in this paper, namely, linear or conic similarity functions, Lukasiewicz t-(co)norm, and $L_1$- or $L_2$-norm distance-from-satisfaction functions, a solution to the SOCP can be found in time $O(|R|^{3.5})$.*

A detailed derivation of the SOCP and proof of the theorem can be found in the extended version of the paper [4].

### 2.5 Weight learning

Weight learning in PSL is performed using standard techniques by optimizing the log-likelihood of interpretation $I$:

$$\log \mathbb{P}(I(\mathcal{G})) = -\delta(V(I), \mathbf{0}) - \log Z \quad (12)$$

Above, $\delta(V(I), \mathbf{0})$ is as in Definition 2. The gradient of Equation (12) with respect to weight $w_k$ depends on the

```
Function: MAP-Inference
1.1  I_0(y) ← all zeros assignment
1.2  R ← all grounded rules activated by I(x) ∪ I_0(y)
1.3  while R has been updated do
1.4      i ← current iteration
1.5      O ← generateConvexProb(R)
1.6      I_i(y) ← optimize(O)
1.7      foreach Proposition y ∈ y do
1.8          if I_i(y) > θ(θ = 0.01) then
1.9              R_y ← activated rules containing y  R ← R ∪ R_y
1.10         end
1.11     end
1.12 end
```
**Algorithm 1**: MAP-Inference in PSL

distance function $\delta$ used. For $\delta(\cdot, \cdot) = \|\cdot, \cdot\|_1$ (which we used in the experiments), the gradient is given by:

$$-\frac{\partial}{\partial w_k} \log \mathbb{P}(I(\mathcal{G})) = \|V_k(I), \mathbf{0}\|_1 - \mathbb{E}(\|V_k(I), \mathbf{0}\|_1),$$

where $\mathbb{E}(\|V_k(I), \mathbf{0}\|_1)$ is the expected value of $\|V_k(I), \mathbf{0}\|_1$ with respect to the currently learned weights. We experimented with two ways of optimizing the above gradient: we used BFGS, a popular quasi-Newton method [18], and the Perceptron algorithm [7], where in both cases the expectation was approximated with the value of $\|V_k(I), \mathbf{0}\|_1$ in the MAP state, which is a frequently used approximation since computing the expectation is intractable.

### 2.6 The PSL System

We implemented PSL in Java using a relational database[1] for storage and efficient retrieval during the rule grounding. Given a set of rules and database handle, the system grounds out all rules that could be potentially unsatisfied and builds the numeric model corresponding to those rules which can be solved using any standard numeric optimization toolbox[2]. If the optimal solution determined by the numeric solver changes the truth value of an atom, the system automatically determines all affected rules and grounds out any rules that might become unsatisfied as a consequence. A number of data structures are maintained to efficiently determine such changes. Changes to the ground rules are reflected in the numeric model which is maintained throughout the reasoning process and updated within the solver. This allows the solver to exploit knowledge about the previous optimal solution to quickly converge on the new solution. The system will be available at http://psl.umiacs.umd.edu.

## 3 Experiments

This section presents an empirical evaluation of PSL that addresses two questions:

---
[1]We used the freely available H2 database (http://www.h2database.com/)
[2]We used MOSEK (http://www.mosek.com).

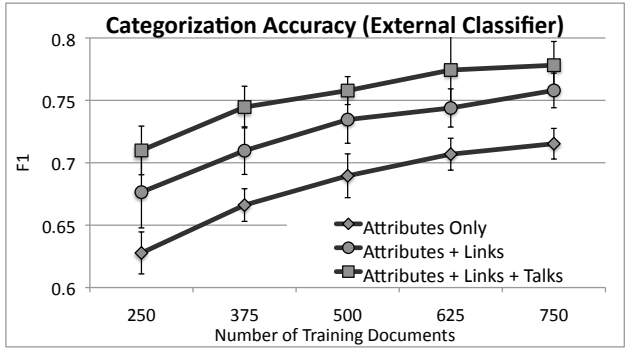

Figure 1: F1 score on classification against number of training documents.

1. Is PSL effective at modeling relational inference tasks?
2. How useful are the novel features provided by PSL?

We study these questions on two distinct problems, namely (a) category prediction and similarity propagation for Wikipedia documents and (b) ontology alignment on a standard corpus of bibliographic ontologies. After describing the data and the experimental methodology, we present results that demonstrate PSL's effectiveness on relational inference. We then investigate in more detail the importance of sets and the benefit of reasoning about similarity.

### 3.1 Wikipedia Category Prediction

We collected all Wikipedia articles that appeared in the featured list[3] in the period Oct. 7-21, 2009, thus obtaining 2460 documents. We used featured articles because they are richly connected, both by their hyperlinks and by their network of human editors [3]. After stemming and stop-word removal, we represented the text of each document as a tf/idf-weighted feature vector. Each document belongs to one of 19 distinct categories, which were obtained by using the category under which each featured article was listed. Some of the original categories that were similar were merged to ensure that each category contains sufficiently many documents. The data contains the relations Link(fromDoc,toDoc), which establishes a hyperlink between two documents; Talk(document,user), which states that the user edited the "Talk" page of the given document;[4] and HasCat(document,category), which states that the document has a particular category. We used the last two years of edits to the talk pages. To reduce noise, we discarded talks that were marked as "minor" by the users themselves or were authored by users with no user names, which typically correspond to automated bots or instances of vandalism. The dataset is available at http://psl.umiacs.umd.edu.

We applied PSL to two distinct tasks in this data set. First,

---
[3]http://en.wikipedia.org/wiki/Wikipedia:Featured_lists

[4]In Wikipedia, each page has an accompanying Talk page where editors discuss potential changes to the content.

to verify that PSL can handle common relational problems, we experimented with a collective classification setting, where relational information is used in an effort to improve over a classifier trained on the tf/idf-weighted word features on a holdout document set. The goal is to predict HasCat for each test document. The second task tests PSL's ability to reason about and propagate similarities. In this task, the text features of the documents are used only to compute a measure of similarity between any given pair of documents and are not used directly as features in a classifier. At test time, the categories of a small subset of the documents, called "seed documents," are observed, and the goal is to *propagate* these assignments to unlabeled documents based on the similarity between them and the relationships in which they participate. The accuracy of the inferred similarities is evaluated through the correctness of the category assignments of the unobserved documents, inferred through the HasCat relation, and thus, on the surface this task may seem almost identical to collective classification. However, we emphasize that in the similarity propagation task the text of documents is not used directly but only to measure similarities between the documents; no model is trained on the textual features of the documents. In other words, there is no assumption that the documents in the test set are from the same domain, or even the same language, as those in the training set. Thus, to perform well, our model needs to effectively propagate these similarities through the relational structure.

The methodology in both tasks is as follows. We randomly split the entire corpus of documents into two equal-sized sets $A$ and $B$ and remove all relations between documents in different sets. We use the data in $A$ for training and then test on set $B$. The results we report in Figures 1 and 2 are averages over 16 independent runs, and the vertical error bars show the standard deviations.

For the collective classification task, we trained a Naive Bayes classifier over the text features of a randomly selected subset of $X$ documents from $A$ (Figure 1 will show results for varying $X$). The predictions of this classifier were provided as evidence through the ClassifyCat(wordFeatures, category) similarity function. We used the remaining documents in $A$ for training the rule weights with the BFGS algorithm. We experimented with three sets of rules. The first set, **Attributes-Only**, is a baseline that uses only the following rule, which simply copies the predictions of the Naive Bayes classifier:

$$\text{ClassifyCat}(\text{A},\text{N}) \overset{\sim}{\Rightarrow} \text{HasCat}(\text{A},\text{N})$$

The second set, **Attributes+Links:**, contains an additional rule stating that hyperlinked documents tend to have the same category:

$$\text{hasCat}(\text{B},\text{C}) \overset{\sim}{\wedge} \text{link}(\text{A},\text{B}) \overset{\sim}{\wedge} \text{A} \neq \text{B} \overset{\sim}{\Rightarrow} \text{HasCat}(\text{A},\text{C})$$

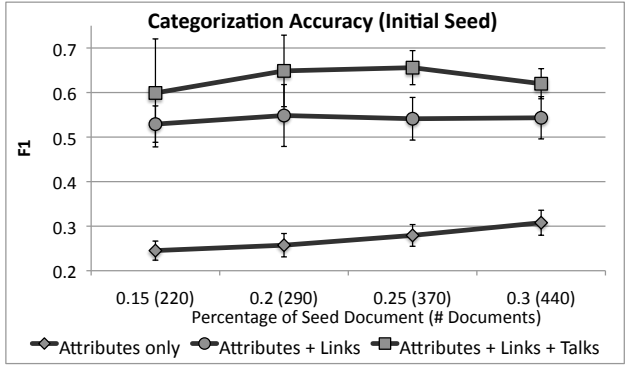

Figure 2: F1 score on category prediction against percentage of seed documents

The third set **Attributes+Links+Talks** contains an additional rule stating that two documents talked about by the same user have the same category:

$$\texttt{talk(D, A)} \;\tilde{\wedge}\; \texttt{talk(E, A)} \;\tilde{\wedge}\; \texttt{hasCat(E, C)} \;\tilde{\wedge}\;$$
$$\texttt{E} \neq \texttt{D} \;\tilde{\Rightarrow}\; \texttt{HasCat(D, C)}$$

Each set of rules includes an additional constraint which ensures that each document can have at most one category. To compute the precision, for each document we select the category to which it is most related, according to the propagated similarities, and compare that category to the ground truth. We note that no document categories were provided during testing, but that Naive Bayes training requires a significant number of labeled documents. The F1 scores (computed as the harmonic mean between precision and recall) on collective classification are shown in Figure 1. All differences are statistically significant at $p = 0.01$. We observe that considering link relationships yields an average 6.5% improvement over the baseline, whereas link and talk relationships combined improve the baseline F1 score by an average of 10.6%. As expected, improvements are larger for smaller training corpus size where the base classifier is less accurate.

On the similarity propagation task, we randomly designate $X\%$ of the documents in the train ($A$) and test ($B$) sets as seed documents and reveal their category during inference. As a baseline (**Attributes-Only**), we use a rule stating that documents with similar word vectors, as measured by cosine similarity, have the same category. As before, we used two additional rule sets by extending the baseline with rules concerning link and talk relationships (**Attributes+Links** and **Attributes+Links+Talks** respectively). We emphasize that in contrast to the first task, here we do not use the words as features in the model but only to compute similarities between documents. Figure 2 shows the average F1 scores for varying percentage of seed documents. All observed differences are significant at $p = 0.01$, except for the two left-most points, where the significance is at $p = 0.02$. We observe that propagating category assignments via link and talk relationships yields a huge improvement over the attribute similarity baseline.

These results demonstrate that the relational structure in the Wikipedia data set is helpful and that PSL can effectively exploit it to model both tasks.

### 3.2 Ontology Alignment

Ontology alignment has received growing attention in recent years, in part due to the explosion of interest in web services, information exchange over the web in general and the semantic web in particular. A large number of approaches have been proposed (see [6] and [10] for surveys). Ontology alignment is a particularly challenging problem due to the complexities of ontologies themselves. Ontologies define concepts, relations, and objects and a host of possible relationships between those basic entities. In addition, ontologies have an associated semantics which constrains feasible alignments to ensure consistency.

Using the general PSL framework, we designed a set of 21 rules and constraints expressing our understanding for how similarity propagates within ontologies. Some of these rules are hard rules, like a rule stating that one concept from ontology $O_1$ can be equivalent to at most one concept in ontology $O_2$, that ensure the consistency of a computed alignment. The majority of the rules are soft-weighted rules, like rules stating that concepts are equivalent if their names or their parents are similar. For example:

$$\texttt{type(A, concepts)} \;\tilde{\wedge}\; \texttt{type(B, concepts)} \;\tilde{\wedge}\; \texttt{name(A, X)}$$
$$\tilde{\wedge}\; \texttt{name(B, Y)} \;\tilde{\wedge}\; \texttt{similarID(X, Y)}$$
$$\tilde{\wedge}\; \texttt{A.source} \neq \texttt{B.source}$$
$$\tilde{\Rightarrow}\; \texttt{similar(A, B)}$$

states that two concepts $A, B$ with similar names defined in different source ontologies are likely to be similar. similarID is a similarity function implemented using a slightly modified Levenshtein metric.

If two concepts align, then it is likely that their respective sets of sub-concepts align as well, which we capture in the following rule using the set equivalence operator $s_{\{\}}$ defined in Section 2.1.

$$\texttt{type(A, concepts)} \;\tilde{\wedge}\; \texttt{type(B, concepts)} \;\tilde{\wedge}\; \texttt{similar(A, B)}$$
$$\tilde{\wedge}\; \texttt{A!} = \texttt{B} \;\tilde{\Rightarrow}\; \{\texttt{A.subclassOf}\} \stackrel{s_{\{\}}}{=} \{\texttt{B.subclassOf}\}$$

A fraction of the rules consider attribute similarity as source of evidence while the remaining rules focus on equivalences of related entities, such as sub-concepts, super-concepts, incident relations, and others. The full set of rules is included in the extended version of this paper [4]. We extended standard string similarity measures, such as Levenshtein and Dice similarity, to measure the similarity between attributes. Given a pair of ontologies, we convert each ontology into a knowledge base of ground atoms and load it into the database.

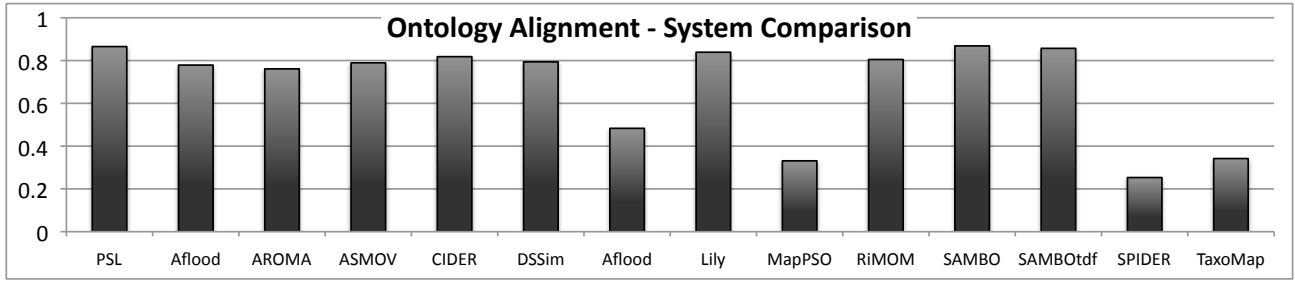

Figure 3: F1 Measure comparison of different ontology alignment systems on real bibliographic ontologies

We do not claim that 21 rules suffice to capture all intricacies of ontologies, but catering toward any of them is as easy as adding more rules or similarity measures. The ease with which rules can be modified and similarity functions integrated into PSL allows model designers to quickly evaluate their intuitions by testing different rules and similarity functions with little implementation effort.

To evaluate the performance of our set of PSL rules, we conducted an experimental study using the OAEI benchmark [5]. The Ontology Alignment Evaluation Initiative (OAEI) invites researchers to compare their ontology alignment systems on a fixed set of benchmark ontology pairs for which reference alignments are provided[5]. We used the same set of rules for all ontology pairs and compared the reference alignment to the alignment inferred by PSL. Since the reference alignments provided by OAEI declare equivalences to be either true or false, we used a threshold of $0.5$ on the inferred similarities.

Figure 3 compares the F1 score of PSL against the reported scores of other systems that participated in the evaluation initiative [5] on the real-world ontologies included in the benchmark (300 level). The ontologies used in our evaluation contained approximately 100 entities (concepts, properties) each. We use one ontology pair for training rule weights and then test on the 3 remaining pairs; we report average F1 over all possible test ontology pairs. For weight learning, we used the Perceptron algorithm. Our PSL model obtains an F1 score of $0.865$, which shows that PSL can learn accurate weights from a single pair of ontologies and generalize to the remaining pairs. We observe that using only a small set of PSL Rules, we achieve alignment results that are comparable to the leading ontology matching systems which have been subject to considerable research and implementation effort.

We also experimented with a methodology closer to the one used in the OAEI initiative by manually fixing weights and testing on all ontology pairs. The results were almost identical to the ones in Fig. 3. However, because the weights were informed by our observations on the learned weights from above, we consider these results "contaminated."

[5] http://oaei.ontologymatching.org/2008/

**Utility of Sets**

In this section we quantify the utility of sets for probabilistic relational reasoning on the ontology alignment task. For this purpose, we explored the behavior of the complete and setFree versions of our ontology alignment PSL program on the 20X ontologies of the OAEI benchmark, which contains ontology pairs with randomly inserted attribute and structural noise. We tried to replicate this suite of benchmark ontologies such that we can control the level of noise. For a given level of attribute noise $a$ and structural noise $s$, we replace a fraction of $a$ attributes with random strings and remove a fraction of $c$ relationships from the ontology. The results closely match the 20X ontology pairs.

The PSL rules we used for ontology alignment above contain set constructs. For instance, one rule states that if two concepts are similar, then the sets of their respective children overlap. To contrast standard PSL to setFree-PSL we created a second set of rules in which all rules using sets were replaced by their setFree counterparts using the conversion scheme outlined in Section 2.3. We learned rule weights on one generated pair of ontologies and then tested on a different, independently generated pair with the same noise levels. The results are averages over 10 independent trials. Figure 4 compares the results for two levels of structural noise, $0.2$ in a) and $0.4$ in b), and with attribute noise varying from $0$ to $0.8$. All differences are statistically significant at $p = 0.01$. We observe that complete PSL consistently outperforms the setFree version, yielding improvements from $9.3\%$ to $57\%$ as attribute noise increases.

### 3.3 Similarity and Scalability

Lastly, we discuss the similarity aspect of PSL in the context of the presented experiments. For one, having a semantics centered around similarity allows PSL to easily incorporate a wide range of similarity measures. In our Wikipedia experiments, we integrated cosine similarity and an existing implementation of Naive Bayes. For ontology alignment, we used previously proposed string similarity measures such as Levenshtein, Dice, and others.

However, because our gold standard evaluation data did not contain similarity values but was in terms of hard truth statements, were were unable to evaluate the quality of the similarity values inferred by PSL. Instead, to achieve comparability to the data at hand, we used some means of similarity post-processing, e.g., in ontology alignment we treated two concepts as exactly aligned if their similarity was greater than 0.5. This raises the question of whether a discrete formulation of the problem within PSL would lead to better performance. To answer this question, we implemented a discrete version of PSL called 0-1 PSL based on mixed integer conic programming which requires all result atoms to be either 0 or 1, i.e. *true* or *false*.

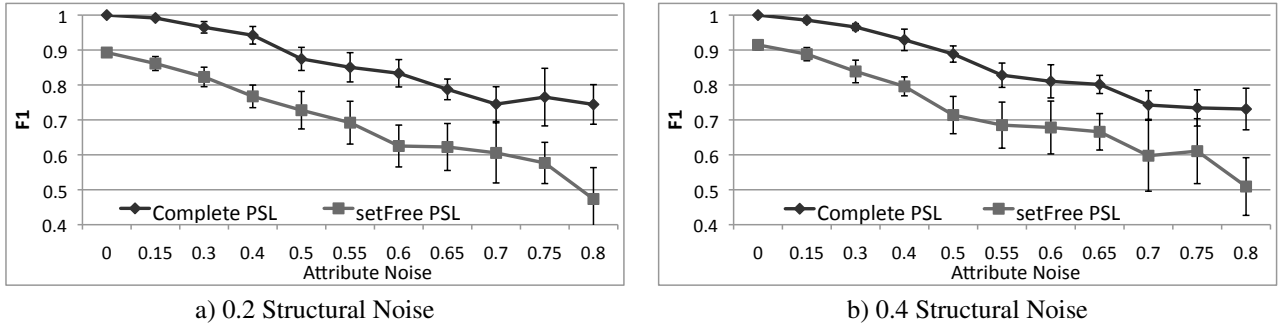

Figure 4: F1 comparison of complete- and setFree PSL on ontology alignment with varying structural and attribute noise

We repeated the Wikipedia category prediction experiments using 0-1 PSL[6]. The measured performance of 0-1 PSL was equal to the continuous formulation (i.e. differences were statistically insignificant) on the similarity propagation task and slightly worse compared to standard PSL (at $p = 0.02$) on collective classification.

While the performance was virtually identical, PSL inference of the discrete formulation took significantly longer. On the similarity propagation task, PSL inference in the original continuous setting took an average of 83 seconds including data loading and preparation. Discrete inference took more than 11 times longer at an average of 974 seconds. Similarly, for collective classification, PSL inference took 18.5 minutes for a complete run including classifier training, whereas 0-1 PSL required almost an hour (54 minutes) on average. Since the ontology alignment task is much smaller, inference times were under 5 seconds with most of the time spend on parsing and data loading. These statistics also demonstrate the efficiency and scalability of standard PSL inference.

## 4 Related Work

PSL builds upon a large body of research in SRL, in which relational structure is parametrized in order to define a probabilistic graphical model over the properties and relations of the entities in a domain, e.g., BLPs [13], PRMs [11], RMNs [21], MLNs [19]. Like these models, PSL also supports probabilistic reasoning over relational structure. However, unlike previous work, PSL additionally supports reasoning about similarities of entities, or sets of entities, and integrates these capabilities into a unified framework. In terms of expressivity, PSL is closest to Hybrid MLNs [22], which allow the use of numeric-valued predicates. However, because Hybrid MLNs were not specifically designed for use with similarities, they do not include support for reasoning about set similarity, and reasoning in them is intractable in general. In contrast, because PSL restricts the numeric-valued predicates it allows to similarity predicates, which evaluate in the interval $[0, 1]$, it can incorporate logical and similarity reasoning in a principled way by using t-(co)norms, thus making use of a well-developed theoretical framework, e.g., [9]. The advantage of this is that, while in Hybrid MLNs numeric and Boolean values are always combined by multiplication, in PSL a variety of truth combining functions may be used. For example, as discussed in Section 2.1, here we used Lukasiewicz t-(co)norms because they lead to sparser linear optimization problems. Support for similarity computations between relation-defined sets is an essential feature of PSL and has not been previously explored in any of the above models. PSL is related to imperative frameworks, such as, IDFs [17] that use a programming language to define structural dependencies. While being very general, such frameworks are more complex to use and require implementation by the user, such as providing the MCMC sampler with custom-made proposal functions for each application studied. With PSL, a user only specifies the model – inference and learning do not require user input. PSL is also related to approaches, such as kFoil [15], that base similarity computation with kernel functions on relational structure. While also treating similarities as distances, kFoil addresses a different problem from the one studied here, by assuming that instances are independent. In contrast, PSL not only uses relational structure as features, but also to propagate similarities. In addition, we point to the large body of work in probabilistic and fuzzy logic programs which shares conceptual similarities with PSL. However, PSL uses a very different probabilistic model which captures cyclic depen-dencies, handles inconsistencies, and enforces domain constraints, among other things.

---

[6]For ontology alignment, we were unable to tune the discrete solver to find an optimal solution, possibly due to the more complex relational structure and wide usage of set constructs.

## 5 Conclusions and Future Work

We introduced a new general framework that integrates probabilistic reasoning about similarity in a relational context and demonstrated its effectiveness on two distinct tasks that involve reasoning about similarity. We experimentally validated the utility of set constructs and similarity inference - two novel features supported by PSL. Directions of future work include studying different distance from satisfaction functions, such as $L_2$ distance and applying PSL to other domains, especially ones in which we can validate PSL on ground truth data with similarity values.

## Acknowledgment

We thank Avi Pfeffer, Kristian Kersting, and the anonymous reviewers for their helpful comments and suggestions. This material is based upon work supported by the National Science Foundation under Grant No. 0937094. LM is supported by a CI Fellowship under NSF Grant No. 0937060 to the Computing Research Association. Any opinions, findings, and conclusions or recommendations expressed in this material are those of the authors and do not necessarily reflect the views of the NSF or the CRA.